%% file: 3671-Sinclair.tex
\documentclass[11pt,a4paper]{article}

\usepackage[acceptedWithA]{tacl2021v1}

\usepackage{times}
\usepackage{latexsym}
\usepackage{graphicx}
\usepackage{url}
\usepackage[T1]{fontenc}

\usepackage{xspace,mfirstuc,tabulary}

\newcommand{\ex}[1]{{\sf #1}}

\newif\iftaclinstructions
\taclinstructionsfalse 
\iftaclinstructions
\renewcommand{\confidential}{}
\renewcommand{\anonsubtext}{(No author info supplied here, for consistency with
TACL-submission anonymization requirements)}
\newcommand{\instr}
\fi

\iftaclpubformat 

\else

\fi

\usepackage[utf8]{inputenc}

\usepackage{microtype}

\usepackage{amsmath}
\usepackage{mathtools}
\usepackage{enumitem}
\usepackage{booktabs}
\usepackage{graphicx}
\usepackage{color}
\usepackage{comment}
\usepackage{multirow}
\usepackage{subcaption}
\usepackage{tikz}
\usepackage{dsfont}
\usepackage{amsmath,amssymb}
\usepackage{amsthm}
\usepackage{float}

\usepackage{tabularx}
\usepackage{colortbl}
\usepackage{multirow}
\usepackage{booktabs}

\usepackage{helvet}
\newcolumntype{L}{l<{\kern\tabcolsep}@{}}
\definecolor{custom_purple}{HTML}{C963C6}

\newtagform{brackets}{[}{]}
\usetagform{brackets}

\usepackage{hyperref}
\newcommand{\newref}[2]{\hyperref[#1]{\ref*{#1}#2}}

\theoremstyle{definition}
\newtheorem{definition}{Definition}[section]

\usepackage{linguex}
\AtBeginDocument{\settowidth{\Exlabelwidth}{(0)}}
\AtBeginDocument{\setlength{\Exlabelsep}{2pt}}
\AtBeginDocument{\setlength{\SubExleftmargin}{10pt}}
\AtBeginDocument{\setlength{\Extopsep}{0.4\baselineskip}}

\newcommand{\raq}[1]{\textcolor{blue}{[RF: #1]}}
\newcommand{\js}[1]{\textcolor{purple}{[JS: #1]}}
\newcommand{\todo}[1]{\textcolor{red}{\textsc{ToDo}}: \textcolor{orange}{#1}}

\newcommand{\edit}[1]{\textcolor{teal}{ #1}}

\newcommand{\new}[1]{\textcolor{black}{ #1}}

\newcommand{\newer}[1]{\textcolor{black}{ #1}}
\newcommand{\splitpg}[1]{\textcolor{red}{ #1}}

\newcommand*\samethanks[1][\value{footnote}]{\footnotemark[#1]}
\newcommand\Mark[1]{\textsuperscript{#1}}
\newcommand\blfootnote[1]{%
  \begingroup
  \renewcommand\thefootnote{}\footnote{#1}%
  \addtocounter{footnote}{-1}%
  \endgroup
}

\title{Structural Persistence in Language Models:\\
Priming as a Window into Abstract Language Representations}

\author{Arabella Sinclair\Mark{1,2}\Thanks{Equal contribution}~~~Jaap Jumelet\Mark{2}\samethanks~~~Willem Zuidema\Mark{2}~~~Raquel Fern\'andez\Mark{2} \\
\begin{tabular}{c c} 
\Mark{1}School of Natural and Computing Sciences & \Mark{2} Institute for Logic, Language and Computation\\
University of Aberdeen & University of Amsterdam\tabularnewline
\end{tabular}\\
{\small\texttt{arabella.sinclair@abdn.ac.uk~\{j.w.d.jumelet|zuidema|raquel.fernandez\}@uva.nl}}
}
  
\begin{document}
\maketitle

\input{sections/abstract}
\input{sections/introduction}

\input{sections/background}

\input{sections/measure}
\input{sections/corpus}
\input{sections/methods}
\input{sections/experiments_core}

\input{sections/experiments_factors}

\input{sections/conclusion}

\section*{Acknowledgements}
We would like to thank the anonymous reviewers for their extensive and thoughtful feedback and suggestions, which greatly improved our work, as well as Chris Brew for his helpful guidance as action editor. We would also like to thank members of the ILLC past and present for their useful comments and feedback, specifically, Dieuwke Hupkes, Mario Giulianelli, Sandro Pezzelle, and Ece Takmaz. Arabella Sinclair worked on this project while affiliated with the University of Amsterdam. The project has received funding from the European Research Council (ERC) under the European Union’s Horizon 2020 research and innovation programme (grant agreement No.\ 819455).

\bibliography{priming}
\bibliographystyle{acl_natbib}




\end{document}

%% file: sections/abstract.tex

\begin{abstract}
We investigate the extent to which modern, neural language models are susceptible to structural priming, the phenomenon whereby the structure of a sentence makes the same structure more probable in a follow-up  sentence. We explore how priming can be used to study the potential of these models to learn abstract structural information, \newer{which is a prerequisite for good performance on tasks that require natural language understanding skills.} 
We introduce a novel metric and release \textsc{Prime-LM}, a large corpus where we control for various linguistic factors which interact with priming strength.
We find that Transformer models indeed show evidence of structural priming,
but also that the generalisations they learned are to some extent modulated by semantic information. Our experiments also show that the representations acquired by the models may not only encode abstract sequential structure but involve certain level of hierarchical syntactic information. More generally, our study shows that the priming paradigm is a useful, additional tool for gaining insights into the capacities of language models and opens the door to future priming-based investigations that probe the model's internal states.\textsuperscript{1} \blfootnote{\textsuperscript{1}\hspace{0pt} Our code and data can be found at \url{https://github.com/dmg-illc/prime-lm}.} 
\end{abstract}

%

%% file: sections/introduction.tex

\section{Introduction}
\label{sec:intro}

It has become increasingly clear that modern, neural language models (LMs) are capable of representing and learning a broad range of linguistic phenomena \citep{Gulordava2018ColorlessHierarchically, DBLP:conf/naacl/HewittM19, tenney2019bert, rogers2020primer, warstadt2020blimp}.
However, many open questions remain about the extent to which specific LMs have indeed acquired specific linguistic constructions, about whether these models encode an abstract notion of \new{structure} in their representations, and about the best ways to even assess the syntactic abilities of these models.
A rich literature has emerged in the last few years addressing these questions, often taking inspiration from methodologies developed in theoretical linguistics, psycholinguistics, neurolinguistics and language acquisition research \citep{futrell2019neural, ettinger2020bert, doi:10.1146/annurev-linguistics-011619-030303, gauthier2020syntaxgym, baroni2021proper},
where the same questions have been asked about the human mind/brain for centuries.

Building on this tradition, this paper turns to \textbf{\new{structural} priming} to investigate the degree to which LMs encode abstract structural information \new{independent from the concrete words that make up sentences.} 
This phenomenon refers to the fact that humans are more likely to produce---or to more easily comprehend---a sentence of a certain
structure $X$
(the \textit{target}) when they have been exposed before to a sentence of a similar structure $X$ (the \textit{prime}), than if they had been prompted with a sentence of a different structure $Y$.
For example, a native speaker of English will be more inclined to produce the target sentence with a prepositional object in \NNext[a] after having read sentence \Next[a] instead of \Next[b], and, vice versa, be more inclined to produce the double-object target sentence \NNext[b] after having read 
\Next[b] instead of \Next[a].
\new{Similar effects are also observed in language comprehension.}

\ex.
\a. \emph{A teacher cooked a chicken for a worker}
\b. \emph{A teacher cooked a worker a chicken}

\ex.
\a. \emph{The guest threw the pot to the lady}
\b. \emph{The guest threw the lady the pot}

\newer{Evidence for structural priming---to the extent that it can be shown to be independent from lexical overlap and other confounds---is taken as evidence for a linguistic structural level of representation that abstracts away from the surface form of sentences. Thus whether or not language models display structural priming can provide insights as to their structural awareness, which is necessary for downstream tasks requiring natural language understanding skills. Previous experiments designed to test structural encoding in LMs are inconclusive. On the one hand, studies on structural probing \citep{DBLP:conf/naacl/HewittM19} and on syntactic evaluation tasks \citep{warstadt2020blimp} have yielded evidence for its presence. On the other hand, other sets of experiments have indicated that current LMs are surprisingly indifferent to word order \citep{hessel-schofield-2021-effective,pham-etal-2021-order,DBLP:conf/emnlp/SinhaJHPWK21} and rely on superficial heuristics when resolving downstream tasks \citep{mccoy-etal-2019-right, sinha-etal-2021-unnatural}. Such unresolved tensions between results --and the active debate about them-- highlights the need for developing additional methodologies that isolates structure from the lexico-semantic cues given to the model. In this paper, we leverage findings from structural priming in human language processing to develop a systematic experimental pipeline with the aim of assessing the extent to which pre-trained neural language models learn representations that encode structural information---a prerequisite for their good performance on natural language understanding tasks. 
}

\new{We use the term `structural priming' \cite{pickering2008structural} rather than `syntactic priming' (first described in Katryn Bock's \emph{Syntactic persistence in language production}, 1986) \nocite{bock1986syntactic}
because 
it comprises priming of abstract structural information that is not restricted 
to syntactic hierarchical rules, such as the linear positions of semantic roles or the sequential order of parts of speech. In this paper, we focus mostly on the latter and touch upon syntactic rules in Section~\ref{sec:complex}.}

In Section~\ref{sec:PEmeasure}, we define an efficient novel metric for measuring the effect of priming. For our experiments, we create \textbf{\textsc{Prime-LM}}, a large-scale corpus for examining structural priming consisting of $\sim$1.3M prime-target sentence pairs, as we describe in Section~\ref{sec:corpus}.
Earlier work on priming in LMs by \citet{prasad-etal-2019-using} operationalised priming as adaptation or implicit learning 
and thus fine-tuned 
the model weights in between prime and target. While our priming effect metric is compatible with priming as adaptation, our experiments in this paper concentrate on priming 
after recent exposure to linguistic context 
without updating the model weights. This allows us to assess the structural representational abilities acquired by the models during training and investigate to what extent such structural information remains active \newer{at inference time.}

\new{In Section~\ref{sec:core} and \ref{sec:factors} we use our corpus and priming paradigm to answer three main research questions: (1)~Are modern neural language models susceptible to structural priming? (2)~Which factors influence the strength of the priming effect? And: (3)~what is the nature of the structural representations acquired by those models?
Our results show that Transformer language models \newer{\textit{do}} exhibit structural priming. \newer{This finding provides evidence that} abstract structural information is encoded by the models to some degree and persists as a model makes predictions about upcoming sentences. The strength of the priming effect is influenced by several factors, including the semantic similarity and the proximity between prime and target, as well as the amount of exposure to a given structure during prompting. Our final experiment moreover reveals that the structural representations encoded by the model may not only be sequential but involve a certain level of hierarchical syntactic structure.}

%% file: sections/background.tex
\section{Background}
\label{sec:background}

\subsection{Structural Priming in Humans}
\label{sec:priming}

Priming is the dominant paradigm in psycholinguistics for investigating the extent to which human language processing involves a level of structural representation independent from other types of linguistic knowledge. The rationale behind this paradigm is that if speakers \new{are sensitive to} 
sentence structure independently from sentence content, then it is reasonable to assume that such structural information is an integral part of the representations built during processing.

\new{In human language processing, structural priming effects are well attested both in comprehension and production}~\cite[][among others]{bock1986syntactic,pickering1998representation,bock2000persistence,pickering2008structural,goldwater2011structural,pickering2013persistent,reitter2014alignment,tooley2014parity}.
Several studies have shown that the strength of the priming effect increases after repeated exposure to a given structure \cite{kaschak2011structural,jaeger2013alignment} and tends to decay if material intervenes between prime and target \cite{reitter2011computational}. 
\new{Other experiments have shown} that ungrammatical and semantically incongruent sentences (e.g., \emph{the waitress brunks the book to the monk}) lead to similar priming effects as well-formed sentences \cite{ivanova2012comprehension,ivanova2017you}, which suggests that structural persistence effects are robust enough in the absence of semantic and lexical cues.

 \new{Yet, structural priming has been found to be affected by various aspects. For example,
priming effects are stronger with lower-frequency than higher-frequency constructions \cite[e.g.,][]{scheepers2003syntactic,bernolet2010does,pickering2013persistent}. Similarly,
some types of lexical repetition between prime and target
have been shown to enhance structural  priming, suggesting that there is a lexical component involved \cite{pickering1998representation,cleland2003use}.
Semantic relatedness between prime and target also has a boosting effect, 
albeit smaller than the lexical repetition boost \cite{cleland2003use,mahowald2016meta}.}

In the present study, we take inspiration from this tradition to investigate the 
priming behaviour of neural language models, which in turn depends on them encoding structural information.
\new{Two (not necessarily exclusive) mechanisms have been proposed to account for structural priming in humans: short-term residual activation of structural information across utterances \cite[e.g.,][]{branigan1999syntactic,wheeldon2003phrase} and long-term adaptation or implicit learning involving changes in the probability of a given structure \cite{BOCK2007437,kaschak2011structural,fine2013evidence}.
Here we focus on the ability of large pre-trained LMs to encode structural information 
\newer{given in the preceding context, similarly to residual activation in humans}.
}

\subsection{Structural Sensitivity of Neural LMs}
\label{sec:interpretability}
The increasing capacities of neural language models in recent years have led to a surge in research into their representation 
of language on a fine-grained linguistic level \citep[i.a.]{alishahi2019analyzing, tenney2019bert, rogers2020primer}.
A common approach to examining \newer{language models} is to consider them as `\textit{psycholinguistic subjects}'; by testing hypotheses derived from psycholinguistics we are able to determine to what extent \newer{language models} process language similarly to humans~\citep{futrell2019neural,ettinger2020bert,davis-van-schijndel-2020-discourse,Lakretz2021MechanismsFH}.

To assess the linguistic knowledge of LMs, a range of tools have been deployed.
For instance, by training auxiliary diagnostic classifiers on top of a model's internal states \citep{hupkes2018visualisation}, we can probe whether these states encode certain linguistic properties such as POS tags \citep{DBLP:conf/iclr/TenneyXCWPMKDBD19}, syntactic dependencies \citep{DBLP:conf/naacl/HewittM19, white_non-linear_2021}, \newer{or constructional information \citep{DBLP:conf/coling/MadabushiRDM20, DBLP:journals/corr/abs-2202-12246}}.
Another common approach is the usage of Targeted Syntactic Evaluations, in which the LM's output probabilities are compared on a minimally different pair of a grammatical and ungrammatical sentence \citep{linzen2016assessing, marvin2018targeted, gauthier2020syntaxgym, hu2020systematic}.
This procedure makes it possible to investigate a model's knowledge of specific linguistic phenomena \new{without probing the model's internal representations}, such as negative polarity items \citep{warstadt2019investigating, jumelet-etal-2021-language}, subject-verb agreement \citep{Gulordava2018ColorlessHierarchically, lakretz2019emergence}, and argument binding \citep{warstadt2020blimp}. 

\newer{
Taken together, results from probing, Targeted Syntactic Evaluations and other existing evaluation paradigms can certainly be viewed as providing converging evidence that modern neural LMs learn 
non-trivial structural, linguistic knowledge and do not just memorise 
fragments of texts from the data and simple sequential dependencies. However, although converging, the evidence is not yet conclusive: each of these evaluation paradigms has also been found to occasionally produce false positives. In probing, for instance, a well-known risk is that probes pick up information represented in the internal states of the language model, but not causally involved in the predictions of the model \citep{voita-titov-2020-information}. In Targeted Syntactic Evaluations, the strength of the evidence depends on the quality of the set of alternative explanations that is considered, which ultimately is a matter of judgements and differs for different linguistic constructions \citep{vamvas-sennrich-2021-limits}. Recent studies have provided new challenges, including studies 
pointing out the indifference of LMs towards word order \citep[][i.a.]{DBLP:conf/emnlp/SinhaJHPWK21}, their reliance on spurious heuristics \citep{DBLP:conf/iclr/LoveringJLP21}, and their difficulty in dealing 
with negation \citep{ettinger2020bert, DBLP:conf/acl/KassnerS20}.}

\newer{
Hence, the debate about the abilities of language models to learn structural information in general, as well as their success in learning certain linguistic constructions specifically, is far from over. The research we present in this paper starts from the observation that structural priming may provide a much needed, complementary methodology 
that, like Targeted Syntactic Evaluations, examines the behaviour of a model, but also, like probing, informs us about the nature of the internal states. 
We will assess a model's representation of a sentence by measuring its consequences in processing the next sentence.
Instead of examining how the model deals with specific syntactic properties within a sentence, such as number agreement, we measure its encoding of abstract structure at the overall sentence level and the consequences this has for upcoming sentences.
In the next section we explain our approach in detail.}

%% file: sections/measure.tex

\section{Measuring Priming}
\label{sec:PEmeasure}

We capture the effects of priming by measuring the difference in log probability of a target sentence $T_\textsc{x}$ given a prime sentence $P_\textsc{x}$ of the same syntactic structure $\textsc{x}$, vs. $T_\textsc{x}$ given $P_\textsc{y}$, a sentence of the exact same semantic and lexical content as $P_\textsc{x}$ but differing in syntactic structure $\textsc{y}$.
We call this metric the \textit{\textbf{Priming Effect} (PE)}:
\begin{equation}\label{eq:pe}
      {\scriptstyle \log}P(T_\textsc{x}|P_\textsc{x}) - {\scriptstyle \log}P(T_\textsc{x}|P_\textsc{y})
\end{equation}
\new{
By measuring priming based on a fixed prime-target pair our method is akin to structural priming in comprehension.
We condition a target sentence on a prime sentence by concatenating them, separated by a period.
The log probability is computed as the sum of token log probabilities of the LM:} 
\begin{equation}
    {\scriptstyle\log}P(T_\textsc{x}|P_\textsc{x}) = \sum_i {\scriptstyle\log}P_{\textsc{lm}}(T_{\textsc{x}_i}|P_\textsc{x}, T_{\textsc{x}_{<i}})
\end{equation}
For example, the Priming Effect of the example in the introduction would be computed as follows:%
\begin{align*}
    {\it PE}_\textsc{po} &= {\scriptstyle \log}P(T_\textsc{po}|P_\textsc{po}) - {\scriptstyle \log}P(T_\textsc{po}|P_\textsc{do})\\
    {\it PE}_\textsc{do} &= {\scriptstyle \log}P(T_\textsc{do}|P_\textsc{do}) - {\scriptstyle \log}P(T_\textsc{do}|P_\textsc{po})
\end{align*}



\noindent  (where $P_\textsc{po}$, $P_\textsc{do}$, $T_\textsc{po}$, $T_\textsc{do}$ denote sentences 1a, 1b, 2a, 2b).
To ensure our estimates of the priming effect are robust,
we incorporate the procedure of \citet{newman-etal-2021-refining} by pairing each target sentence in a corpus with 10 different prime sentences.

\begin{definition}[\textit{Priming Effect (PE)}]
Measures the effect of priming as the difference in log probabilities:%
\begin{equation*}
\frac{1}{|\mathcal{P}|} \sum_{P_\textsc{x} \in \mathcal{P}(T_\textsc{x})}
  \left[{\scriptstyle \log}P(T_\textsc{x}|P_\textsc{x}) - {\scriptstyle \log}P(T_\textsc{x}|P_\textsc{y}) \right]
    \end{equation*}
\end{definition}

\noindent
where $\mathcal{P}(T_\textsc{x})$ denotes the set of prime sentences that can be matched with target $T_\textsc{x}$.
In our experiments, we report the mean of this metric, taken over large-scale corpora of semantically diverse sentences.

Our Priming Effect method is related to various other metrics that are used in the context of priming and statistics in general.
When the conditional probabilities are close to 0, -- as is the case for our corpora with a mean sentence probability around $10^{-18}$ -- this metric approaches the log odds ratio that is used by \citet{mahowald2016meta}.
This allows our scores to be directly comparable to their results on human priming.
A more general connection can be made between our metric and Bayes factors \citep{jeffreys1961theory, kass1995bayes}, which determine the strength of evidence
and are, similar to our metric, also defined as a log probability ratio.

\new{
\citet{prasad-etal-2019-using} model priming as an implicit learning procedure \citep{chang2000structural}, instantiated as a fine-tuning-based adaptation process \citep{van-schijndel-linzen-2018-neural}.
The adaptation effect is then obtained by comparing the impact of a single prime structure on two target sentences of opposing structure, comparing their perplexity before and after fine-tuning:}
\[{\it PP}(T_\textsc{x}) - {\it PP}(T_\textsc{x}|P_\textsc{x}) > {\it PP}(T_\textsc{y}) - {\it PP}(T_\textsc{y}|P_\textsc{x})\]
\new{
The authors also identify a problem: this metric is proportional to the prior perplexities ${\it PP}(T_\textsc{x})$ and ${\it PP}(T_\textsc{y})$. They resolve the issue by regressing out this relationship. This procedure, however, is based on  assumptions that do not always hold, namely, that the relationship between the priming metric and the prior perplexities of the two targets is linear and homoscedastic.
In our experiments we found neither assumption to hold empirically, and hence we opted to directly compare the impact of two prime sentences on a single target sentence.
This way we do not need to regress out confounding effects of prior probabilities, since we are comparing the same quantity (the target sentence) to two primes.
The contrast between these metrics is illustrated by the diagrams in Figure~\ref{fig:measure}.}

\new{Note that our Priming Effect metric could be applied to the priming-as-adaptation paradigm as well, by comparing the target sentence probabilities of two fine-tuned models.
In the experiments presented in this paper, we focus on priming as residual activation and thus do not update the model weights, which makes the approach more computationally efficient.}

\begin{figure}[!h]
\begin{center}
\begin{tikzpicture}[scale=0.1]
    \tikzstyle{every node}+=[inner sep=0pt]
    \draw [black] (45.7,-26.1) circle (3);
    \draw (45.7,-26.1) node {$P_\textsc{x}$};
    \draw [black] (56.4,-20.6) circle (3);
    \draw (56.4,-20.6) node {$T_\textsc{x}$};
    \draw [black] (56.4,-31.3) circle (3);
    \draw (56.4,-31.3) node {$T_\textsc{y}$};
    \draw [black] (16.9,-20.6) circle (3);
    \draw (16.9,-20.6) node {$P_\textsc{x}$};
    \draw [black] (16.9,-31.3) circle (3);
    \draw (16.9,-31.3) node {$P_\textsc{y}$};
    \draw [black] (27.7,-26.1) circle (3);
    \draw (27.7,-26.1) node {$T_\textsc{x}$};
    \draw (53,-39.1) node {\citet{prasad-etal-2019-using}};
    \draw (19,-39.1) node {\textit{Priming Effect} (Eq.~\ref{eq:pe})};
    \draw [black] (48.37,-24.73) -- (53.73,-21.97);
    \fill [black] (53.73,-21.97) -- (52.79,-21.89) -- (53.25,-22.78);
    \draw [black] (48.4,-27.41) -- (53.7,-29.99);
    \fill [black] (53.7,-29.99) -- (53.2,-29.19) -- (52.76,-30.09);
    \draw [black] (19.57,-21.96) -- (25.03,-24.74);
    \fill [black] (25.03,-24.74) -- (24.54,-23.93) -- (24.09,-24.82);
    \draw [black] (19.6,-30) -- (25,-27.4);
    \fill [black] (25,-27.4) -- (24.06,-27.3) -- (24.49,-28.2);
\end{tikzpicture}
\end{center}
\caption{\new{Our Priming Effect metric compares the impact of two prime sentences with different structures on a single target exhibiting one of the structures. \citet{prasad-etal-2019-using} examine the impact of a single prime structure on two target sentences.}}
\label{fig:measure}
\end{figure}

%% file: sections/corpus.tex
\section{The \textsc{Prime-LM} Corpus}
\label{sec:corpus}

We create a large-scale set of corpora designed to examine the priming behaviour of LMs.

\subsection{Syntactic Alternations}
\label{sec:constructions}

In the current experiments, we focus on two types of syntactic alternations, \emph{dative} and \emph{transitive}, which allow for the same content to be expressed by two different structures. The dative alternation includes ditransitive verbs whose complements can be expressed by a double-object (\textsc{do}) structure or a prepositional-object (\textsc{po}) structure (e.g., \emph{the boss gave the dog a bone} vs.\ \emph{gave a bone to the dog}). The transitive alternation includes transitive verbs within an active (\textsc{act}) or a passive (\textsc{pass}) structure (e.g., \emph{the actor followed the student} vs.\ \emph{the student was followed by the actor}).

\ex. \ {\bf Dative}\\
\setlength\tabcolsep{1.5pt}
\begin{tabular}{rl}
\textsc{do}:& \new{\texttt{Dt} \texttt{N}$_{agent}$ \texttt{V} \ \texttt{Dt} \texttt{N}$_{recipient}$ \ \texttt{Dt} \texttt{N}$_{patient}$} \\[-5pt]
\textsc{po}:& \new{\texttt{Dt} \texttt{N}$_{agent}$ \texttt{V} \ \texttt{Dt} \texttt{N}$_{patient}$ \ \texttt{Pr} \texttt{Dt} \texttt{N}$_{recipient}$}
\end{tabular}
\label{ex:d-template}

\ex. \ {\bf Transitive}\\
\setlength\tabcolsep{1.5pt}
\begin{tabular}{rl}
\textsc{act}:& \new{\texttt{Dt} \texttt{N}$_{agent}$ \texttt{V} \ \texttt{Dt} \texttt{N}$_{patient}$}\\[-5pt]
\textsc{pass}:& \new{\texttt{Dt} \texttt{N}$_{patient}$ \  \texttt{Aux} \texttt{V} \ by \texttt{Dt} \texttt{N}$_{agent}$}
\end{tabular}
\label{ex:t-template}

%
%

\noindent
In the transitive case, the active structure is dominant in English~\citep{bock1986syntactic,merriam1989webster}. The proportion of use between structures for the dative alternation is less marked, with different studies showing 
a preference for the direct-object structure~\citep[e.g.,][]{bock1986syntactic,bresnan2007predicting}.

\subsection{Corpus Construction}
\label{sec:corpus_general}
We construct a set of corpora by filling in the templates in \ref{ex:d-template} and \ref{ex:t-template} above. For the content words \new{(nouns and verbs)}, we exploit the vocabulary present in the University of South Florida (USF) free association norms dataset~\cite{nelson2004university}, which contains pairs of cue-target words with their association strength.\footnote{Corresponding to the percentage of human participants who produced the target word when asked to come up with words related to the cue (\url{http://w3.usf.edu/FreeAssociation/}).
}
This allows us to control for the degree of semantic association between prime and target sentences. 
To minimise any effects stemming from word frequency factors, we only include USF content words which appear in the top 5000 most common words according to the COCA corpus~\cite{AMUDUW2015}.

We identify transitive and ditransitive verbs using vocabulary lists targeted at English language learners,\footnote{\url{http://www.aprendeinglesenleganes.com/resources}, \url{https://englishpost.org/transitive-verbs-list}, and \url{https://www.cse.unsw.edu.au/\~billw/ditransitive.html}
}
\new{keeping those that are present in USF and meet the frequency constraints (around 80 verbs in total).}
The ditransitive verbs were manually labelled for the preposition to be used in the \textsc{po} structure (\emph{to/for}) and the transitive verbs were annotated with their past participle form to be used in the passive construction.
In addition, all verbs were manually labelled for \new{some of the noun categories they can take as arguments (e.g., the transitive verb \emph{wash} was annotated as accepting agents of category \texttt{person} and patients of category \texttt{person} or \texttt{object})}.
Following the same frequency constraints, a set of nouns fulfilling these categories was selected from USF   using the WordNet closure categories of \textit{person}, \textit{social\_group, social\_control, institution}, \textit{physical\_entity}, and \textit{object}, which we further hand split into \textit{non-edible}, \textit{edible}, and \textit{drinkable}.\footnote{To ensure compatibility with the indefinite article \emph{a/an} (see section~\ref{sec:corpus_core}), uncountable nouns were discarded.} This yielded 119 nouns in total.

From this vocabulary, we are able to generate many realisations of our sentence templates through sampling,
respecting the grammaticality of the sentences produced.
\new{Three native speakers of English manually examined a subset of sentences for each verb and syntactic alternation to confirm that the sentences produced are well formed. This resulted in the elimination of a few ditransitive verbs for which the \textsc{do} structure was considered awkward. The final corpus contains 48 transitive and 16  ditransitive verbs.}


\new{Using this template-based method,} we create a series of corpora that satisfy various semantic and lexical constraints. For each of these corpora we specify a corpus size of 15,000 prime-target pairs per syntactic target structure (\textsc{do}, \textsc{po}, \textsc{act}, \textsc{pass}), which are obtained by pairing 1,500 different target sentences with 10 semantically different primes.\footnote{The corpus size of 15,000 was determined based on Cochran's Formula for sample size determination \citep{CochranWilliamG1977St}, with a \textit{p}-value and margin of error of 0.01.}
Overall, \textsc{Prime-LM} contains $\sim$1.3M prime-target pairs.


\subsection{The \emph{Core} Corpus}
\label{sec:corpus_core}

\new{\textsc{Prime-LM} consists of a \emph{core} corpus and a set of variants over this core.}
In the core corpus, we ensure that prime and target sentences (1) include different determiners, either  \textit{a/an} or \textit{the}, (2) do not share any nouns nor verbs, and (3) only contain nouns and verbs that are not semantically associated across prime and target according to the USF free association norms dataset.\footnote{The average cosine similarity across pairs of words in prime and target computed with \texttt{word2vec} embeddings by \citet{fares-etal-2017-word} is 0.2 for both nouns and verbs.}
For the \textsc{po} structure, we additionally make sure that prime and target differ in preposition (\textit{to} vs.\ \textit{for}), which makes all the prime and target sentences in the dative alternation lexically fully disjoint.
For the transitive alternation, this is not possible since the preposition \emph{by} must appear in the \textsc{pass} structure.
Other than that, we completely limit lexical overlap for transitive constructions by using alternate auxiliary verb forms (\textit{is} vs. \textit{was}) for the passive prime and target, and create their active counterparts by using the corresponding tense of the auxiliary to maintain semantic equivalence.
All sentences in the dative alternation are in the past simple tense.

\new{As an illustration, below we show two examples from the \emph{core} corpus following the scheme in Figure~\ref{fig:measure}, where $P$ are the prime sentences and $T$ the target:}

\ex.
$P_{\textsc{po}}$: \emph{A pilot bought a pie for an attorney}\\
$P_{\textsc{do}}$: \emph{A pilot bought an attorney a pie}\\
$T_{\textsc{po}}$: \emph{The professor sent the tea to the band}

\ex.
$P_{\textsc{act}}$: \ \emph{The nurse purchased the beer}\\
$P_{\textsc{pass}}$: \emph{The beer was purchased by the nurse}\\
$T_{\textsc{pass}}$: \emph{An engine is wrapped by a colonel}

We create different variants of the core corpus that isolate specific aspects shown to influence \new{structural} priming in human sentence processing. They are described in Section~\ref{sec:factors} together with the corresponding experiments. Example sentences for each of our corpora can be found in Table~\ref{tab:corpus_examples}.


%% file: sections/methods.tex

\section{Language Models}
\label{sec:models}
We focus our experiments on the class of \textit{auto-regressive} LMs,\footnote{Also known as \textit{causal} or \textit{left-to-right} language models, predicting the probability of the next token solely on prior context.} which are trained to predict the next token, in line with human incremental language processing.
\new{
Our methodology can be applied to masked LMs as well; we briefly reflect on this in the discussion (\S\ref{sec:discussion_conclusion}).}
The main focus of our analysis is directed upon on Transformer models \citep{DBLP:conf/nips/VaswaniSPUJGKP17}, that constitute the current state of the art in language modelling, and have been shown to produce representations that correlate strongly with human brain signals \citep{Schrimpf2020.06.26.174482}.
This is the set of models we consider: 
\begin{itemize}[itemsep=1pt,leftmargin=13pt]
    \item
\textbf{GPT2}, in its four sizes \citep[\textsc{small, medium, large, xl;}][]{radfordlanguage}, and its \textit{distilled} version \citep{DBLP:journals/corr/abs-1910-01108};
\item
  \textbf{DialoGPT}, three GPT2 models of increasing size that have been fine-tuned on dialogue data \citep{zhang-etal-2020-dialogpt};
\item
  \textbf{GPT-Neo} in three sizes \citep[\textsc{125m}, \textsc{1.3b}, \textsc{2.7b};][]{gpt-neo}, which is based on GPT3 \citep{DBLP:journals/corr/abs-2005-14165}.
\end{itemize}

\noindent
All Transformer LMs are imported with the \texttt{transformers} library \citep{wolf-etal-2020-transformers}.
The extraction of the model probabilities is done using the \texttt{diagNNose} library \citep{jumelet-2020-diagnnose}, which provides support for efficient activation extraction.
Our implementation allows our priming procedure to be efficiently tested on any kind of language model and to be easily reproducible. All our code and corpora will be publicly released upon publication of the paper.

\paragraph{Why should LMs exhibit structural priming?}
\new{Since structural repetition is present in human language use and common in corpora \cite{dubey2008probabilistic}, LMs have, in theory, the potential to learn such structural dependencies during training.
It is, however, not reasonable to expect that models which have been trained on shuffled sentences will exhibit priming, because such models will not be able to adequately carry over a linguistic signal (structural or otherwise) from the prime sentence to the
target.}\footnote{In our experiments we had initially incorporated two LSTM LMs \citep{DBLP:journals/corr/JozefowiczVSSW16, Gulordava2018ColorlessHierarchically}, and indeed due to their shuffled training corpus we did not observe any notable Priming Effect. We are not aware of any available LSTM LM trained on unshuffled data.}
\newer{As mentioned in the introduction and in Section~\ref{sec:interpretability}, }several studies 
have suggested that structural information is being encoded by large language models; yet, other studies showing that LMs are often insensitive to permutations in word order \cite[e.g.,][]{kodner-gupta-2020-overestimation, sinha-etal-2021-unnatural} cast doubt on these results.  
Thus, while there is potential for LMs pre-trained on unshuffled data to encode structural dependencies that are detectable with our priming paradigm, whether they will in fact do so remains an open question, since the language modelling objective (next word prediction) contains no explicit cues for structural information. 
This is precisely the question we address in this work.

\paragraph{Priming behaviour}
\new{To interpret our results we distinguish between three types of behaviour: i)~\textit{symmetrical priming} occurs when a model obtains positive Priming Effects for both constructions within an alternation: the model has fully picked up on the structural congruence between prime and target; ii)~\textit{asymmetrical priming} occurs when a model obtains a positive Priming Effect for one construction, and a Priming Effect close to zero for its counterpart;\footnote{Such asymmetries are common in humans \citep{bock1986syntactic,gries2005syntactic,segaert2016unifying}.}
and iii)~\textit{biased priming} occurs when a model obtains a positive Priming Effect for one construction, but a  negative Priming Effect for its counterpart.
A priming bias indicates that a prime of the preferred structure is more likely to boost any subsequent target that we consider,
regardless of its structural congruence with the prime.
Hence, we take symmetrical and, to some extent, asymmetrical priming behaviour to
represent evidence for the structural priming effect we are interested in.\footnote{This is analogical to, for example, subject-verb agreement: a model that always prefers a plural verb, regardless of the subject number, can't be said to understand the task. A model that scores 100\% on plural verb prediction, but randomly for singular verbs, has an \textit{asymmetric} understanding of the task.}
}

%% file: sections/experiments_core.tex
\section{Core Priming Results across LMs}
\label{sec:core}

\begin{figure}[t]
    \centering
    \includegraphics[width=\columnwidth,trim=.25cm .2cm .2cm .2cm,clip]{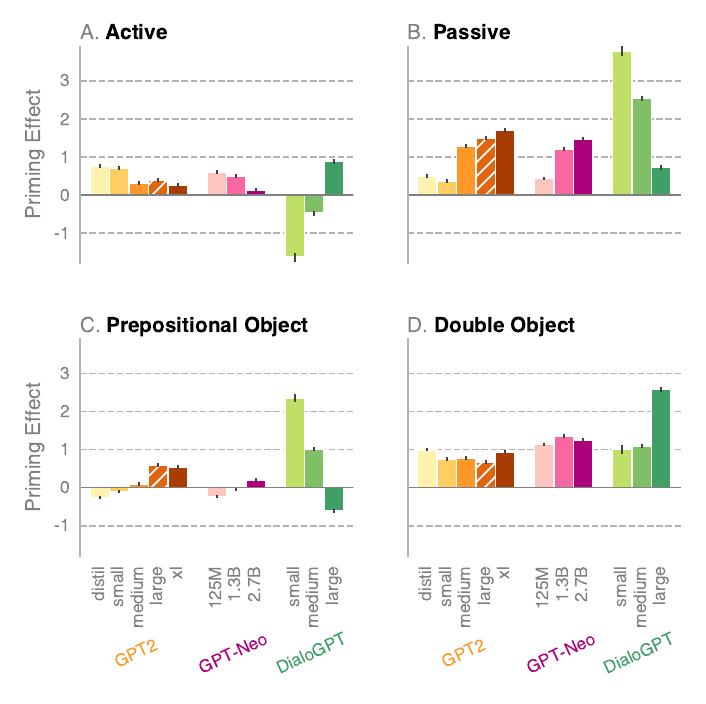}
    \caption{\textit{Priming Effect} results of all models on the \textit{core} corpus, across the four syntactic structures.
    Error bars denote 99\% confidence intervals of the mean.
    The GPT2-\textsc{large} model that will be explored in more detail in \S\ref{sec:factors} has been highlighted.
    }
    \label{fig:core_results}
\end{figure}

We initially test all LMs described in the previous section on our \emph{core} corpus, designed to control for lexical overlap and semantic similarity. This provides a \new{clean 
experimental setup, where the only element shared between prime and target is the abstract sequential structure.} 
The results are reported in Figure \ref{fig:core_results}, split by the structure type of the target sentence.
It can be seen that across many models a positive Priming Effect is present.
We will now discuss these results in more detail.


\new{There are two models that exhibit symmetrical priming for both transitive and dative alternations: GPT2-\textsc{large} and GPT2-\textsc{xl}.
The other GPT2 models exhibit symmetrical priming for transitive as well, but exhibit moderate asymmetrical priming behaviour for dative, with priming occurring only for double-object structure.
DialoGPT-\textsc{small} exhibits biased priming for transitive constructions: a negative Priming Effect on active constructions, but a large positive Priming Effect for passive constructions.
This shows that for this particular model a passive prime boosts the probability of an active target more than an active prime does, resulting in a negative effect.}

\paragraph{Model size}
We can investigate the impact of model size by comparing the results of the different sizes of the models we consider.\footnote{Note that the different sizes of a model are trained on the same amount of data; only the number of parameters is affected.} 
Larger models may have more potential for encoding finer-grained structural information \cite[see e.g.,][]{hu2020systematic}.
If model size were to have a positive effect on structural priming this might manifest itself in two ways: either (1) the Priming Effect increases for both structural alternatives, or (2) the priming bias towards one structure decreases. 
We do not see evidence of (1). 
As for (2) regarding bias, results differ between transitive and dative.
For the GPT2 models the asymmetrical priming towards double objects is decreased, resulting in symmetrical priming for both GPT2-\textsc{large} and GPT2-\textsc{xl}.
For the DialoGPT results on transitive we can see that the severe bias towards passive decreases as model size is increased, resulting in symmetrical priming behaviour for DialoGPT-\textsc{large}.
For dative constructions, however, the larger model size gives rise to a priming bias towards double objects: in this case increasing model size actually has a detrimental effect on the model's priming behaviour.
From this we conclude that sensitivity to structural priming is partly driven by model size, but is likely to depend on a more intricate combination of factors related to model architecture and training data, which needs to be investigated further in future work.

\paragraph{Best model}
The models that exhibit more susceptibility to structural priming across all four construction types are GPT2-\textsc{large} and GPT-2-\textsc{xl}.
For GPT2-\textsc{large} the congruent conditional probability $P(T_\textsc{x}|P_\textsc{x})$ was larger than the incongruent one $P(T_\textsc{x}|P_\textsc{y})$ 60.5\% of the time for active, 81.0\% for passive, 65.4\% for prepositional object, and 72.1\% for double object.
In the subsequent experiments we will focus our analysis on GPT2-\textsc{large} and use more specialised experimental conditions within the priming paradigm to dig deeper into the potential of the model for encoding structural information.

%% file: sections/experiments_factors.tex

\begin{table*}[ht!]
  \small
  \setlength{\abovecaptionskip}{0pt}
    \centering
    \include{tables/corpus_examples_actpass}

    \caption{%
    Example sentences for the core corpus and each condition described in \S\ref{sec:lexical_dependence}, \S\ref{sec:anomalous} and \S\ref{sec:complex}. The same manipulations illustrated here for the \textsc{act} and \textsc{pass} also hold for the dative alternation.
    }
    \label{tab:corpus_examples}
\end{table*}

\section{Impact of Specific Factors}
\label{sec:factors}

The next battery of experiments isolates various factors that have been shown to be of influence to priming in human language processing.
For each experimental condition, we present a specialised corpus followed by an analysis of the priming effects exhibited by GPT2-\textsc{large} on this data, comparing them to the model's behaviour on the core corpus. Examples from the core and specialised conditions can be found in Table~\ref{tab:corpus_examples}.

\subsection{Lexical Dependence}
\label{sec:lexical_dependence}
In the \emph{core} corpus,  prime and target sentences are semantically unrelated, which ensures that priming effects cannot stem from the model assigning higher probabilities to words that are similar or identical to those present in the prime sentence. 
In the following two experiments we relax this constraint to investigate the extent to which lexical semantic similarity and lexical repetition across prime and target have an impact on structural
priming effects.

\subsubsection{Semantic Similarity}\label{sec:semsim}
We create versions of the core corpus where prime and target sentences have different degrees of lexical semantic similarity. Concretely, a pair of words sharing the same semantic role in the prime and target is considered semantically similar if they (a) are associated according to the USF norms, and (b) have a cosine similarity  \cite[computed with embeddings from][]{fares-etal-2017-word} equal or higher than the 90\%-percentile of the distribution of similarities in the core corpus.\footnote{This results in a cosine similarity threshold of $\sim$0.4.}

\new{In human experiments, semantic similarity has been found to boost priming~\citep{goldwater2011structural}, both in nouns \citep{cleland2003use}, and in verbs~\citep{pickering1998representation}.}
We isolate the effects of verb and noun similarity by creating conditions where (1) only the verb, (2) all nouns, or (3) all content words are semantically similar across prime and target sentences. These additional constraints result in a more limited set of possible sentence pairs for condition (3), and thus in a reduced corpus of 228 (transitive) and 1648 (dative) prime-target pairs rather then 15,000.\footnote{In this case, to maximise the number of unique pairs, we allow a varying number of primes to target, rather than observing the 10-to-1 prime-target setup of the other corpora.}

\begin{figure*}
    \centering
    \includegraphics[width=0.96\textwidth]{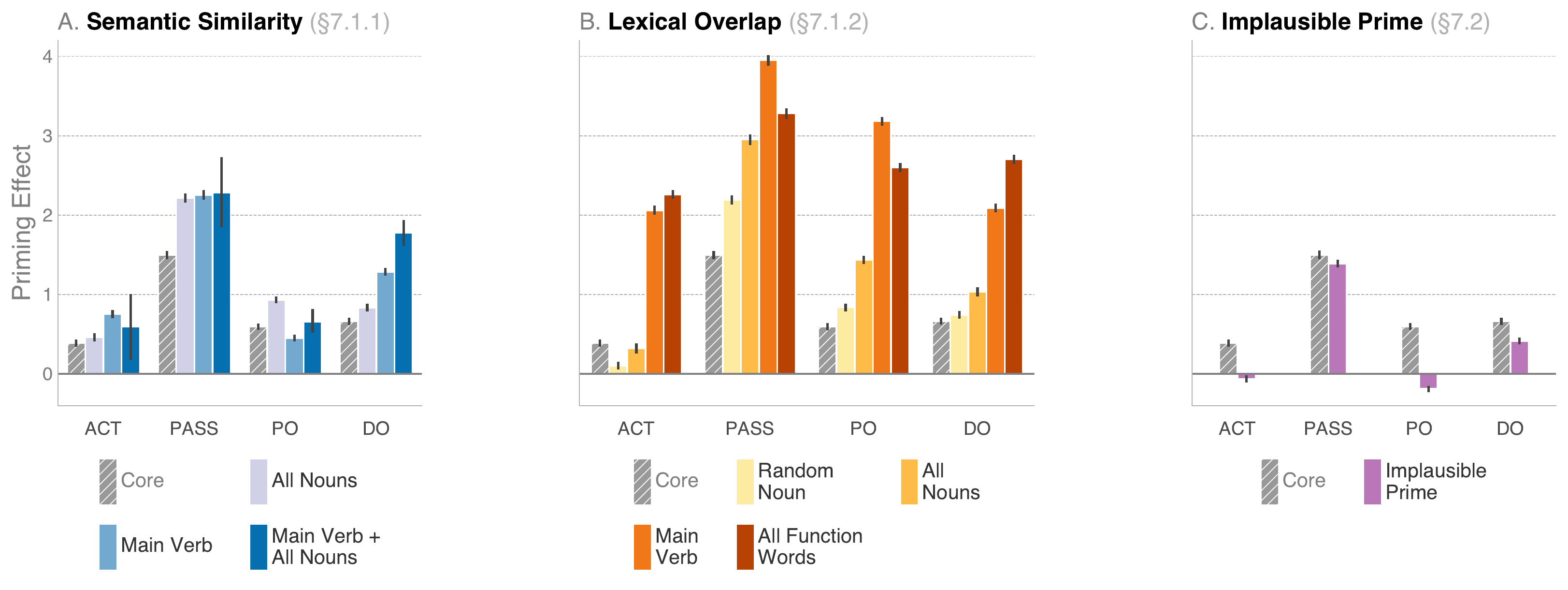}
    \caption{
    Results for GPT2-\textsc{large} on the experiments described in and \S\ref{sec:lexical_dependence} and \S\ref{sec:anomalous}: \textbf{A.} measures the impact of semantic similarity between prime and target, \textbf{B.} the impact of lexical overlap between prime and target, and \textbf{C.}  whether priming is affected by the semantic implausibility of the prime.
    }
    \label{fig:experiments_abc}
\end{figure*}

\paragraph{Results}
We find greater Priming Effect across constructions in this setup compared to the core corpus, although this is less pronounced for the \textsc{po} structure.
As can be seen in Figure~\newref{fig:experiments_abc}{A}, a semantically similar verb in prime and target leads to an increase of the Priming Effect, comparable to the condition where all nouns are similar.  With the exception of \textsc{do}, we do not observe an additive effect: when all content words are similar, the Priming Effect is not substantially higher than when only the verb is similar.

\subsubsection{Lexical Overlap}
Lexical overlap between prime and target in the core corpus was avoided in both content and function words. Here we systematically introduce lexical repetition across prime and target sentences.
We create versions of the core corpus where lexical overlap takes place with respect to only (1) one of the nouns at random but with the same semantic role across prime and target (\emph{agent, patient, recipient}, see \S\ref{sec:constructions}), (2) all nouns, (3) the verb, and (4) all function words (i.e., any determiners, prepositions, and auxiliary verbs are shared across prime and target, without content words being shared). 



\paragraph{Results}
As can be seen in Figure~\newref{fig:experiments_abc}{B}, overall the presence of lexical overlap greatly boosts \new{structural}  priming effects.
For all constructions, verb overlap leads to higher priming effects than repeating one noun or even all nouns.
Surprisingly, overlap of function words has the
highest boosting effect for \textsc{act} and \textsc{do}.\footnote{\new{This contrasts with psycholinguistic evidence suggesting that structural priming is not led by function-word priming in humans \cite{bock1989closed,tree1999building}.}}
To place these results into context, we calculate the Priming Effect when prime and target are identical sentences.
\new{Language models are known to fall prone to repeatedly generating the same sentence \citep{foster-white-2007-avoiding, DBLP:conf/aaai/FuLSS21}; hence this value can be considered a ceiling.}
We obtain a PE of 2.5 for \textsc{act}, 7.2 for \textsc{pass}, 9.2 for \textsc{po}, and 10.1 for \textsc{do} constructions. None of the lexical overlap conditions we consider reaches the corresponding ceiling.
\subsection{Semantic Implausibility} 
\label{sec:anomalous}
\new{In this experiment, we test whether the effects found in the core corpus are robust to manipulations concerned with the semantic plausibility of the sentences used as stimuli. This helps to diagnose to what extent any structural information encoded by the model is autonomous from semantics. To this end,}
we construct a version of the corpus where the prime sentences are specifically designed to be semantically implausible. \citet{Gulordava2018ColorlessHierarchically} employed a similar method in their study of long-distance agreement dependencies, finding that RNN's ability to predict number agreement was robust to nonsensical sentences. The authors interpret this result as evidence that the networks track abstract 
structure, in line with Chomsky's (\citeyear{chomsky1957}) proposal that grammaticality is distinct from meaningfulness in the human language faculty.
Here we further test this hypothesis by analysing whether the LM is susceptible to \new{structural}  priming effects when the prime sentence is nonsensical. As mentioned in \S\ref{sec:priming}, humans do exhibit structural priming effects when prompted with incongruent sentences \cite{ivanova2012comprehension,ivanova2017you}.
We construct semantically implausible primes via sampling nouns at random among noun categories that do not respect the verb selectional restrictions.  This results in grammatically correct, yet nonsensical sentences such as \textit{`the iron threw the hero to the chocolate'}. \new{The same constraints regarding absence of semantic similarity and lexical overlap between prime and target present in the core corpus apply here as well.}

\paragraph{Results}
The results of this experiment are shown in Figure~\newref{fig:experiments_abc}{C}. We find here that the Priming Effect exhibits 
asymmetrical priming behaviour,
indicating that the prime structure itself is more likely to boost any subsequent target regardless of shared structural properties.
The Priming Effect disappears and becomes negative for the \textsc{act} and \textsc{po} constructions, while for \textsc{pass} and \textsc{do} it decreases when compared to the results on the core corpus, but remains positive.
While some degree of abstract structural information present in the nonsensical sentences may be exploited to predict the target construction, the asymmetrical behaviour suggests that structural encoding is not fully independent from semantic plausibility.



\subsection{Activation Strength}
\label{sec:activation_strength}

In the following two experiments, we test whether \new{structural} priming effects are affected by \new{the proximity of prime to target and by increased exposure to the priming structure.}
We maintain 
the strict setting of our core corpus, where prime and target are semantically and lexically unrelated, thus testing to what extent the activation of abstract \new{structural information across sentences} is affected by recency and cumulativity factors. 

\begin{figure}
    \centering
    \includegraphics[width=\columnwidth]{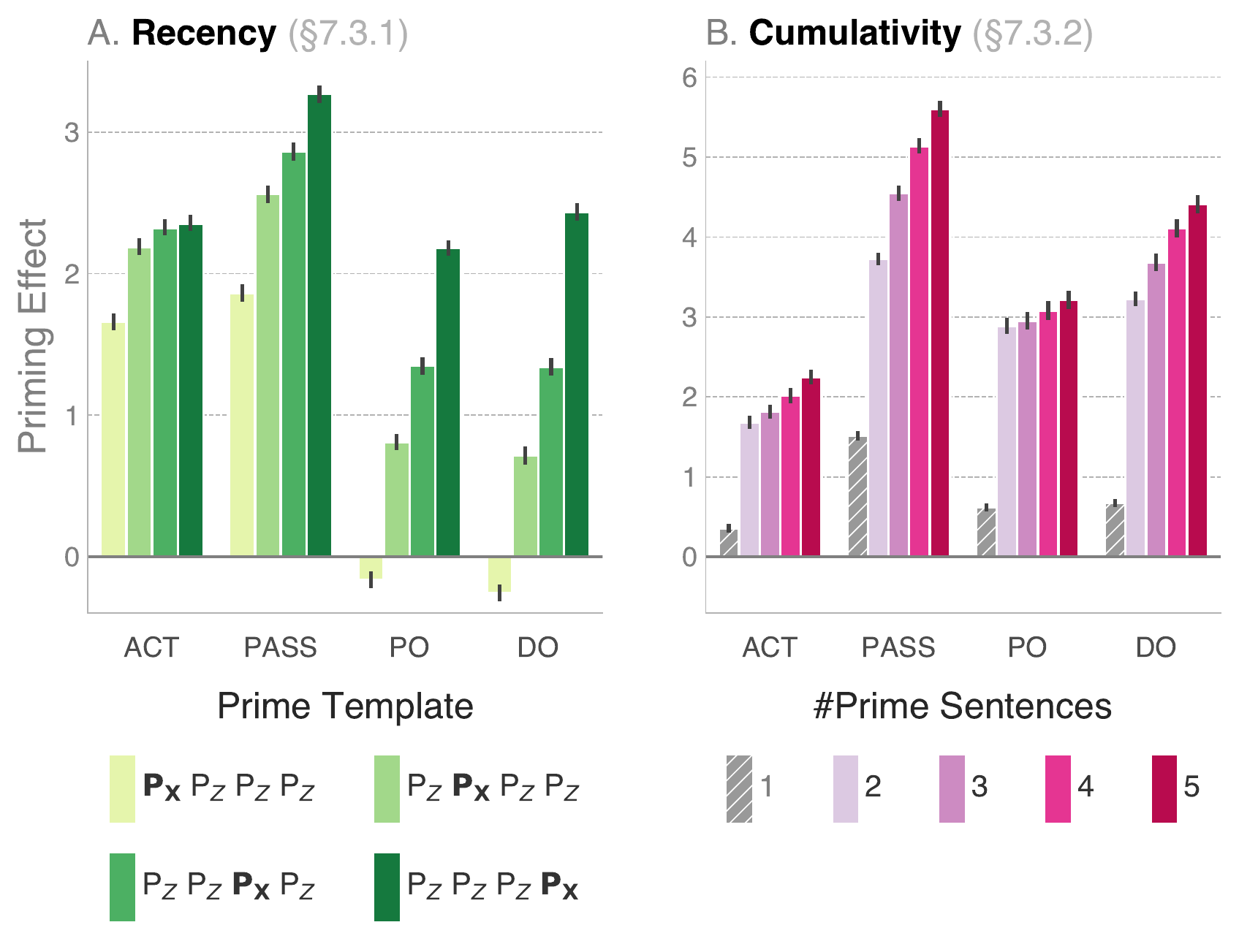}
    \caption{
    Results for GPT2-\textsc{large} on the experiments described in \S\ref{sec:activation_strength}:
    \textbf{A.} recency effect on priming, by increasing the distance between prime and target with additional intransitive sentences: each bar denotes a different position of the prime ($P_\textsc{x}$), surrounded by intervening sentences ($P_\textsc{z}$);
    \textbf{B.} cumulative effects on priming, by increasing the number of prime sentences before a target.
    }
    \label{fig:cumulative_recency}
\end{figure}

\subsubsection{Recency}
To vary the proximity of prime to target, we create a set of \emph{padding} sentences, using intransitive verbs, personal pronouns, and different auxiliary verbs to those used in our core corpus, including modal auxiliary verbs (e.g., \textit{you might come, he did remain, they should appear}). These sentences were designed to contain frequent vocabulary with no lexical overlap nor semantic similarity to the prime and target sentences in the core corpus.
A context in this setting consists of a sequence of 4 sentences, within which the priming sentence will vary in position relative to the target.
This setup ensures that any priming observed is not influenced by the total length of the context, but solely by the position of the prime.
In this condition, the Priming Effect is computed as follows:
\begin{equation}
{\scriptstyle \log}P(T_\textsc{x}|P_\textsc{z}^*~P_\textsc{x}~P_\textsc{z}^*) - {\scriptstyle \log}P(T_\textsc{x}|P_\textsc{y})
\end{equation}
where $P_\textsc{z}$ denotes the sequence of intransitive padding sentences.

\paragraph{Results}
The results of this experiment are shown in Figure~\newref{fig:cumulative_recency}{A}, which shows that
increasing the \new{proximity} between prime and target has a highly \new{positive} impact on the strength of priming. 
\new{
Interestingly, the PE for the transitive cases is still relatively high even when the distance between prime and target is at its largest, whereas for the dative cases the PE has dropped drastically. 
This may indicate that the syntactic configuration of a transitive sentence is not corrupted as much by the intermediate intransitive sentence as the configuration of a dative sentence.
}

\subsubsection{Cumulativity}

\new{To investigate the effect of cumulativity,} we create a version of the core corpus where for each target sentence we sample multiple primes and concatenate them, resulting in priming contexts which vary between 1 and 5 sentences in length. 
All prime sentences in the prompt satisfy the semantic constraints with respect to the target that were outlined in \S\ref{sec:corpus}.
In this case, the Priming Effect is measured as follows:
\begin{equation}
{\scriptstyle \log}P(T_\textsc{x}|P_\textsc{x}^+) - {\scriptstyle \log}P(T_\textsc{x}|P_\textsc{y})
\end{equation}
in other words, the Priming Effect of a sequence of congruent primes $P_\textsc{x}^+$ is expressed with relation to the log probability of a \textit{single} incongruent prime sentence $P_\textsc{y}$. 

\paragraph{Results}
As shown in Figure~\newref{fig:cumulative_recency}{B}, for all constructions the Priming Effect increases monotonically as the number of congruent prime sentences increases.
\new{This resonates with the potential of large LMs for few-shot learning: the multiple priming sentences appear to act as ``demonstrations'' \cite[in the sense of][]{DBLP:journals/corr/abs-2005-14165} of a given structure, which presumably increases the activation of that type of structural information.}
This result is a \new{yet} another indication of \new{structural} information being encoded by the model \new{and remaining active across sentences, as the main feature that is repeated across the multiple primes is the shared abstract structure.}

\subsection{\new{Structural Complexity}}
\label{sec:complex}

\new{Finally, we test whether the priming effects present in the core corpus are robust to different degrees of structural complexity between prime and target. In our core corpus, congruent prime and target sentences are constructed from the same sequence of  parts of speech (see \S\ref{sec:constructions}). Results by \citet{reitter2007against} suggest that, for humans, short-term priming via residual activation is better explained by assuming hierarchical representations.
In this experiment, we test whether the structural information encoded by the model is limited to sequential abstract structure or rather involves hierarchical syntactic representations.} 


To gain more insight on the nature of the structural information represented by the model,
we construct a version of the corpus where some of the noun phrases are more complex than simply ``\texttt{Dt}~\texttt{N}''  (e.g., \emph{the awful tea from Spain}). 
The rationale behind this manipulation is the following: if the structure of a sentence is represented in terms of something akin to a hierarchical phrase-structure rule such as \emph{VP $\rightarrow$ NP NP} or \emph{VP $\rightarrow$ NP PP} \new{rather than as a sequence of part-of-speech categories,} then it should not matter whether prime and target differ with respect to the internal structure of the sub-constituents -- we should observe a similar degree of priming whether the noun phrases are complex or not.  Evidence suggests that this is indeed the case for humans \cite{tree1999building,pickering1998representation,branigan2006role}.

\new{We create a version of the core corpus where the noun phrases may contain a prenominal adjective, a prepositional phrase, neither or both in order to introduce varying degrees of complexity. We use a total of 164 adjectives manually labelled for compatibility with the different noun categories. The prepositional phrases are constructed with either \textit{with} or \textit{from}. For the \textit{with} case, we select a set of 27 suitable nouns within the WordNet categories of \textit{clothing, device} or \textit{container}. This results in noun phrases such as ``\texttt{Dt}\texttt{(A)}\texttt{N} \emph{with} \texttt{Dt}\texttt{(A)}\texttt{N}''. For the \textit{from} case, we use 23 country names, resulting in noun phrases such ``\texttt{Dt}\texttt{(A)}\texttt{N} \textit{from} \texttt{N}''. All the additional vocabulary adheres to the same selection procedure as in \S\ref{sec:corpus}, with prime and target being semantically unrelated. 
We test three conditions: (1)~only the prime sentence has a complex NP, (2)~only the target sentence does, (3)~both prime and target have a complex NP -- ensuring different NP structures across prime and target. In all three settings, any semantic role (\emph{agent, patient}, or \emph{recipient}) can be modified to become complex and there is at most one complex NP per sentence.}

\begin{figure}
    \centering
    \includegraphics[width=0.72\columnwidth]{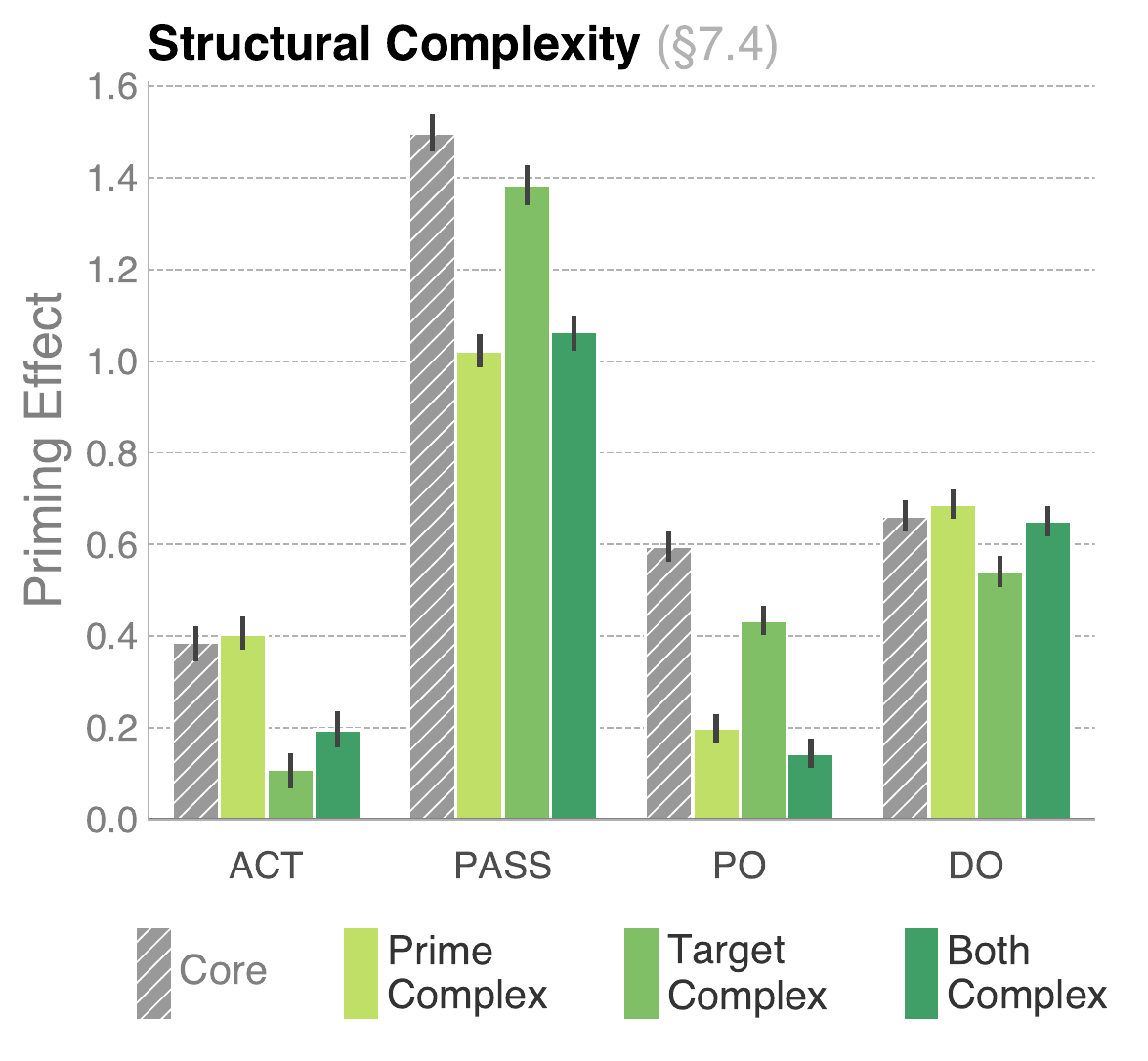}
    \caption{
    Results for GPT2-\textsc{large} on the experiment described in \S\ref{sec:complex}, measuring the impact of increasing the complexity of \new{one noun phrase per sentence in prime and target}.
    }
    \label{fig:complex_np}
\end{figure}

\paragraph{Results}
\new{The results are shown in Figure~\ref{fig:complex_np}.
The first thing to note is that the presence of noun phrases of varying complexity across prime and target 
does not cancel out the Priming Effect: in all cases, the effect remains positive, although there is a decrease for several conditions. We also observe \textit{asymmetrical} priming effects. For example, for transitive with complex prime, e.g., active is  unaffected, whereas the Priming Effect for passive \new{is clearly reduced}.
This suggests that some of the effects observed on the core corpus may be driven by the consistently simple \new{sequential} structures present in that data. Yet, the fact that the priming effect remains positive suggests that there is some degree of hierarchical structural information commonly encoded for both simple and complex NPs, which is carried over to influence the prediction of the target.}

%% file: tables/corpus_examples_actpass.tex
{\sf\scriptsize%
\begin{tabularx}{\textwidth}{@{}llXX@{}}
    {Corpus} & {Condition} & {Prime} (\textsc{act}) & {Target} (\textsc{pass}) \\\arrayrulecolor{gray}\hline
    \textbf{Core} & \textcolor{gray}{---} & \textit{The boy judged the adult.} & \textit{A cousin is forgotten by a secretary.}\\[10pt]
    \multirow{3}{*}{\shortstack[l]{\textbf{Semantic}\\\textbf{Similarity}}} & \textcolor{gray}{Verb Only} & \textit{The chief \textbf{struck} the mayor.} & \textit{A bishop was \textbf{beaten} by a hero.}\\
    & \textcolor{gray}{All Nouns} & \textit{An \textbf{actor} broke a \textbf{glass}.} & \textit{The \textbf{bottle} was wrapped by the \textbf{actress}.}\\
    & \textcolor{gray}{All Words} & \textit{The \textbf{student} \textbf{drank} the \textbf{wine}.} & \textit{A \textbf{beer} was \textbf{prepared} by a \textbf{professor}.}\\[10pt]

    \multirow{4}{*}{\shortstack[l]{\textbf{Lexical}\\\textbf{Overlap}}} & \textcolor{gray}{Random Noun} & \textit{The girl smelled the \textbf{chicken}.} & \textit{A \textbf{chicken} was prepared by a pilot.}\\
    & \textcolor{gray}{Main Verb} & \textit{A woman \textbf{used} a computer.} & \textit{The iron was \textbf{used} by the father.}\\
    & \textcolor{gray}{Function Words} & \textit{\textbf{The} soldier wanted \textbf{the} pie.} & \textit{\textbf{The} book was  carried by \textbf{the} manager.}\\
    & \textcolor{gray}{All Nouns} & \textit{The \textbf{king} smelled the \textbf{wine}.} & \textit{A \textbf{wine} was drunk by a \textbf{king}.}\\ [10pt]

    \multirow{-1}{*}{\shortstack[l]{\textbf{Implausible}\\ \textbf{Prime}}} & \textcolor{gray}{---} & \textit{The \textbf{newspaper grabbed} the \textbf{pot}.} & \textit{A key is removed by an attorney.}\\[10pt]
    & \textcolor{gray}{Prime Complex} & \textit{A lady \textbf{with a red bag} chased a minister.} & \textit{The juice was purchased by the child.} \\ 
    & \textcolor{gray}{Target Complex} & \textit{The physician judged the leader.} & \textit{A \textbf{rich} school was embraced by a business.} \\
    \multirow{-3}{*}{\shortstack[l]{\textbf{Structural}\\ \textbf{Complexity}}} & \textcolor{gray}{Both Complex} & \textit{The \textbf{bad} adult \textbf{with the hat} raised the knife.} & \textit{A son was helped by an author \textbf{from Cuba}.}\\[2pt]
    \arrayrulecolor{gray}\hline
\end{tabularx}
}

%% file: sections/conclusion.tex
\section{Discussion and Conclusions}\label{sec:discussion_conclusion}

In this paper, we investigated three main questions: (1)~Are modern neural LMs susceptible to structural priming?~(2)~Which factors influence the strength of the priming effect?~And:~(3)~what is the nature of the structural representations acquired by those models?~To answer these questions, we designed a series of carefully curated large-scale corpora, proposed a metric to measure the degree to which a model is susceptible to priming,   
and ran a series of 
experiments on several Transformer LMs. This methodology constitutes a new way of assessing the representational abilities of LMs via examining their behaviour in controlled setups, which complements tools like Targeted Syntactic Evaluations and the adaptation-based priming measure by \citet{prasad-etal-2019-using}.


Our results in Section~\ref{sec:core} showed that on our \emph{core} corpus, where we control for lexical overlap and semantic similarity between prime and target,  \emph{most} of the language models we test exhibit \emph{some} degree of priming for \emph{most} of the constructions we study.
This is important, as it opens up the possibility of using priming to investigate what influences the learned representations of these language models. 

In Section~\ref{sec:factors}, we focused on GPT2-\textsc{large} to conduct a series of subsequent experiments to dig deeper into the impact of different factors on the model's susceptibility to priming. In line with psycholinguistic accounts of residual activation, we found that the effects of priming decrease with the distance between prime and target and increase with the amount of exposure to a given structure. Our results indicate that the structural information being encoded is not fully autonomous from semantics: the Priming Effect is highly boosted by semantic similarity and lexical overlap between the words used in prime and target. Such boosting effects are well known to be present in human language processing as well. Furthermore, the Priming Effect partly disappears with semantically implausible prime sentences, suggesting that semantic plausibility is an important cue for the encoding of structure, arguably more so than in human language processing. Finally, we showed that priming effects remain positive in the presence of phrases with differing degrees of complexity across prime and target. 
This offers some insight into the nature of the representations learned by the model: it suggests that, in addition to abstract sequential structure, some degree of hierarchical syntactic information is being represented. 

The current work does not reveal, for the various conditions tested, what the mechanics of the boosting or suppressing effects are. For example, we do not know whether the boosts from lexical overlap or semantic similarity are the result of an improved match with the same structural representations, or 
of independent factors that influence priming behaviour. Similarly, the precise interplay between semantic plausibility and structural encoding remains unclear.
Overall, the pattern of results calls for further investigation using interpretability methods, such as probing and feature attributions, which we plan to pursue in future work.

\begin{figure}[t]
    \centering
    \includegraphics[width=\columnwidth,trim=.25cm .2cm .2cm .2cm,clip]{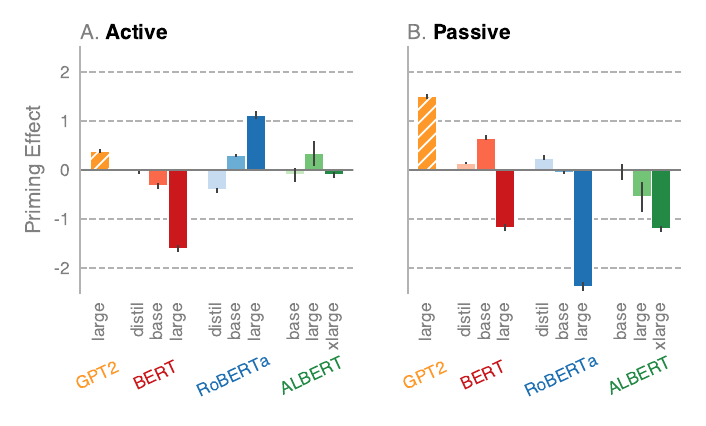}
    \caption{Priming Effects for three masked language models on the \textit{core} corpus: BERT \citep{devlin-etal-2019-bert}, RoBERTa \citep{DBLP:journals/corr/abs-1907-11692}, and ALBERT \citep{DBLP:conf/iclr/LanCGGSS20}.
    To compute sentence probabilities we utilise the pseudo-log-likelihood of \citet{salazar-etal-2020-masked}, masking out one token at a time.
    Results for dative yield a similar pattern.
    }
    \label{fig:mlm_results}
\end{figure}

An additional aspect that requires further study is the role of the training data and its statistics, for example regarding the frequency of the different constructions under investigation and the impact this may have on priming asymmetries within an alternation, and on priming behaviour more generally. 
An important future step to disentangle the factors that may give rise to priming behaviour would involve training a range of different model types on the same data.
This way it becomes possible to interpret the role that model architecture, model size, training objective, and corpus statistics play in shaping the behaviour of the model.
An important class of models to include in such studies are Masked Language Models.
We conducted a preliminary experiment on three such models, which resulted in biased priming behaviour for all (see Figure~\ref{fig:mlm_results}).
We postulate that these models may rely less on the structure of a prime because their bi-directional nature allows them to take the entire target sentence into account.
However, in order to adequately determine that this is entirely due to their training objective, and not due to external factors stemming from corpus statistics, future work could control for this with newly trained models.

Our study reveals novel details about the potential of LMs to represent structural information and the persistence of this information when making predictions about upcoming sentences. But more generally, we believe our findings also demonstrate the usefulness of the priming paradigm for investigating such questions. Even more generally, they illustrate the benefits of repurposing experimental paradigms from psycholinguistics to investigate the knowledge acquired by large neural language models. In that sense, the current paper complements exciting recent work that borrows other paradigms from linguistics and psycholinguistics, including grammaticality judgments, few shot learning, and cloze tests \citep{gauthier2020syntaxgym, DBLP:journals/corr/abs-2005-14165, baroni2021proper, DBLP:conf/iclr/LoveringJLP21}. That is, while syntactic priming offers one window into abstract language representations in neural language models, linguistics offers a whole row of windows that are starting to reveal an exciting vista.